\documentclass{article}
\usepackage{cite}
\usepackage{a4wide}
\usepackage{amsfonts}
\usepackage{amsmath}
\usepackage{algorithm}
\usepackage{algpseudocode}
\usepackage{caption}
\usepackage{subcaption}
\usepackage{booktabs}  
\usepackage{graphicx}
\usepackage{multirow}
\usepackage{listings}
\usepackage{url}
\usepackage[colorinlistoftodos]{todonotes}

\begin{document}

\title{A Python Library for Exploratory Data Analysis on Twitter Data based on Tokens and Aggregated Origin-Destination Information}

\author{Mario Graff$^{1,3,4}$ \and Daniela Moctezuma$^{1,2}$ \and Sabino Miranda-Jiménez$^{1,3}$ \and Eric S. Tellez$^{1,3}$}
\date{%
$^1$INFOTEC Centro de Investigaci\'on e Innovaci\'on en Tecnolog\'ias
de la Informaci\'on y Comunicaci\'on, Circuito Tecnopolo Sur No 112, Fracc. Tecnopolo Pocitos II, Aguascalientes 20313, M\'exico\\
$^2$CentroGEO Centro de Investigaci\'on en Ciencias de Informaci\'on Geoespacial,
Circuito Tecnopolo Norte No. 117, Col. Tecnopolo Pocitos II, C.P., Aguascalientes, Ags 20313 M\'exico\\
$^3$CONACyT Consejo Nacional de Ciencia y Tecnolog\'ia,
Direcci\'on de Cátedras, Insurgentes Sur 1582, Cr\'edito Constructor, Ciudad de M\'exico 03940 M\'exico~\\ ~\\
$^4$Colgate University, Department of Computer Science, 13 Oak Drive, Hamilton, N.Y. 13346 U.S.A ~\\
This work has been accepted to the Special Issue on Data and Information Services for Interdisciplinary Research and Applications in Earth Science of the Computers \& Geosciences Journal}

\maketitle
\begin{abstract}
Twitter is perhaps the social media more amenable for research. It requires only a few steps to obtain information, and there are plenty of libraries that can help in this regard. Nonetheless, knowing whether a particular event is expressed on Twitter is a challenging task that requires a considerable collection of tweets. This proposal aims to facilitate, to a researcher interested, the process of mining events on Twitter by opening a collection of processed information taken from Twitter since December 2015. The events could be related to natural disasters, health issues, and people's mobility, among other studies that can be pursued with the library proposed. Different applications are presented in this contribution to illustrate the library's capabilities: an exploratory analysis of the topics discovered in tweets, a study on similarity among dialects of the Spanish language, and a mobility report on different countries. In summary, the Python library presented is applied to different domains and retrieves a plethora of information in terms of frequencies by day of words and bi-grams of words for Arabic, English, Spanish, and Russian languages. As well as mobility information related to the number of travels among locations for more than 200 countries or territories.
\end{abstract}

\maketitle

\section{Introduction}
\label{sec:introduction}

Twitter is a well-known social network where users let others know their opinions, feelings, and other information, all these in a succinct way. Users of the social network are both real people and organizations connected due to their particular interests. These properties, and its worldwide popularity, make Twitter a useful data source for many research communities interested in capturing people's opinions. Due to users' culture of openness, most messages are indeed public. Researchers can access these messages through Twitter's API; please note that several requirements are needed to access, but those are easier to comply with than other social networks. On Twitter, cyclic events such as weekends, Christmas, New Year, Valentine's Day, among others, naturally emerge by looking only at the tweets' frequency.
For this reason, Twitter reflects in some ways social behavior, and its information is high-correlated with real-world~\cite{Ribeiro2012TrafficTwitter}. Furthermore, people use social media to share and acquire information on any events gives an important clue into social media data's potential. Nonetheless, the overwhelming amount of information generated daily can hide other relevant events, and a specific query is needed to retrieve them. In general, it is difficult to determine the strategy for Twitter's data retrieval related to a social event or phenomenon.

Social networks are also a source of information during a humanitarian crisis. For example, Facebook Disaster Maps\footnote{\url{https://dataforgood.fb.com}} (\cite{Facebook2019FacebookRecovery}) provides access to dynamic maps that include location information, cell site connectivity, and phone battery charging. Social media, particularly Twitter, has been used to obtain situational information (see \cite{Li2020CharacterizingWeibo}), which is a post that can help people make a decision or obtain information in an emergency, see \cite{Vieweg2012SituationalOf}. Another use is the development of a surveillance system, e.g., to detect an outbreak of avian influenza, as described in \cite{Yousefinaghani2019TheStudy}. Social network data has also been used to assess the damages of earthquakes, see \cite{Priya2020TAQE:Disasters,Mendoza2019NowcastingTwitter,Priya2019IdentifyingLearning}. Social media data is useful in several research problems;  \cite{Imran2015ProcessingSurvey} surveys social media usage for analysis of emergencies.

We know that there are a variety of open-source libraries, but many of them do not include a large amount of data resources. Also, it is sometimes necessary to use many libraries to deal with data processing, specifically text processing. In this case, our contribution is based on the \emph{text models} library, which is helpful to both tasks: on the one hand, it provides vocabulary and location data over several years, beginning in December 2015; on the other hand, our library also simplify preprocessing of retrieved data.
Therefore, this contribution presents an open-source Python library, called \emph{text models},\footnote{Repository available at \url{https://github.com/INGEOTEC/text_models}} that aims to help researchers and enthusiasts access aggregated Twitter messages, simplifying an event's data acquisition and processing. More detailed, we collected messages from the Arabic, English, Russian, and Spanish languages, some of them since December 2015. Our motivation to collect these languages is twofold, to cover distinct language families collecting one representative:  Germanic (English), Slavic (Russian), Semitic (Arab), and Romance (Spanish) families; and to consider most spoken languages, according to The Ethnologue 200\footnote{\url{https://www.ethnologue.com/guides/ethnologue200}}, English (1st), Spanish (4th), Standard Arabic (6th), and Russian (8th). In particular, we use the public Twitter stream to listen to tweets with and without geotagged information. We process these messages to produce per-day and per-country aggregated information such as vocabulary usage and user's mobility, among other information. We hope this information converts into practical knowledge allowing a better understanding of social phenomena with resonance on Twitter. Furthermore, we present several usage scenarios (sketching the code used to produce the analysis in the appendix) in this manuscript to demonstrate the potential use of the library's embedded data. These examples make use of retrieved text and mobility information.

The rest of the manuscript is organized as follows. Section~\ref{sec:relatedwork} describes several types of research using Twitter data.
Section~\ref{sec:library} describes the proposed library, and Section~\ref{sec:data-acquisition} presents an overview of the data collected from four languages over the globe. Several analysis' examples of text and mobility data are presented in Section~\ref{sec:analysis}. Finally, Section~\ref{sec:conclusions} is dedicated to summarizing and concluding this manuscript.

\section{Related work}
\label{sec:relatedwork}
Twitter and Facebook have been used as data sources showing that social media can be used to examine disaster management, mobility, public health, politics, and pandemic situations.

For instance, in~\cite{Facebook2019FacebookRecovery}, a disaster map with Facebook's data is presented. These maps are related to the Facebook population, movement, power availability, network coverage, and displacement. With these maps, the authors measure the population as the difference in the number of users based on pre-crisis levels to observe what areas are more affected by the crisis. For instance, it registers those pairs of places where users move and those where users visit to charge their mobile phones. Moreover, the ability of network covering is another aspect recorded by Facebook Data.
Events like a pandemic can also be analyzed using social media data; in~\cite{Li2020CharacterizingWeibo} characterize seven types of situational information regarding the COVID-19 disease. This work uses Weibo data, the primary micro-blogging site in China. As situational information was considered, those related to 1) caution and advice; 2) notifications and measures been taken; 3) donations; 4) emotional support; 5) help-seeking; 6) doubt casting and criticizing; and 7) counter-rumor. Authors learned a model from 3000 COVID-19 related posts manually labeled into these seven categories. The models achieved accuracy performances of 0.54, 0.45, and 0.65, using Support Vector Machine, Näive Bayes, and Random Forest classifiers.
Traffic phenomena are also analyzed using Twitter data. Ribeiro {\em et al.}~\cite{Ribeiro2012TrafficTwitter} proposed a method to identify traffic events on Twitter, geocode them, and display them on a web platform in real-time. The method employs exact and approximate string matching to traffic event identification; for the string matching, the authors manually listed the most frequent terms used for traffic situations.
Human mobility is another activity analyzed using Twitter data~\cite{Jurdak2015UnderstandingTwitter}. Jurdak {\em et al.} propose a proxy for human mobility using geotagged tweets in Australia. 
The population's mobility patterns were analyzed using a dataset of more than 7 million tweets from 156,607 users. The authors compared the patterns found using Twitter data with other technology types, such as call data records. McNeill {\em et al.} \cite{Mcneill2017EstimatingData} proposed to estimate local commuting patterns using geotagged tweets. They found that Twitter's information is a good proxy for inferring the local commuting patterns; even the bias imposed by Twitter users' demographics is not significant. Additionally, natural disaster management, such as earthquakes, could also be analyzed using Twitter information. \cite{Mendoza2019NowcastingTwitter} propose a method based on tracking users' comments about earthquakes to generate a variant of the Mercalli scale computed with Twitter data.\footnote{Qualitative measure of the perceived intensity of an earthquake regarding damages, see \cite{Mendoza2019NowcastingTwitter}.}
Authors use counties (municipalities) as the unit of analysis through several features acquired at that level, such as number of tweets, average words, question marks, exclamation marks, number of hashtags, the occurrence of the \textit{earthquake} word, municipality population, among others.

The event detection task is devoted to discovering phenomena using users' messages (tweets). This task can be performed even without prior knowledge of a particular event. For instance, Kanwar {\em et al.} in~\cite{Kanwar2016DiscoveringTwitter} classify a set of popular events into different categories with more than 5 million tweets and more than 14,000 users in 10 months.
Liu {\em et al.} ~\cite{Liu2016DiscoveringMedia} make a step forward, trying to discover the core semantics from events in social media.
Another example is presented by \cite{Roberts2017UsingEvaluation}, where Twitter's potential as a data source for analyzing urban green spaces is tackled.
Twitter has also been used for detecting and verifying real-time news events. Liu {\em et al.}  in~\cite{Liu2016ReutersTwitter} propose a system for detecting news events and assessing their veracity on Twitter.

\subsection*{Our contribution}
The literature on using social media information to identify and analyze real-world events is vast. Under this context, our contribution is a library and an extensive dataset of multi-year and multi-language preprocessed and aggregated data obtained from people using social networks. The goal is to provide practical tools to simplify and support the analysis of real-world events, as captured in the Twitter platform.

Our \emph{text models} library is confined by the use of text and geolocation data. The analysis of images, videos, and audio is beyond the scope of our work.

\section{Library description}
\label{sec:library}
This section describes all components of our \emph{text models} library and dataset. Our library is a Python package installable from major package managers. The data summarizes data collected from several years, i.e., billions of messages were used to produce the related time series of textual and mobility information.
We used the Tweepy package for collecting data from the Twitter public stream API, see Section \ref{sec:data-acquisition}. Once the library retrieves the data, the data can be post-processed for cleaning purposes, see Section \ref{sec:cleaning}. Section \ref{sec:pmobility} is devoted to geolocation management of our contribution.  Later, in Section \ref{sec:analysis}, some examples of using this library are explained and analyzed. Finally, in Appendix \ref{sec:appendix}, a comprehensive user guide is presented.

\subsection{Data Acquisition}
\label{sec:data-acquisition}

For all languages, we use the Tweepy package to download public tweets\footnote{Official site of the Tweepy package, available at \url{https://www.tweepy.org/}}. There are two ways for this download process: one to obtain tweets without geotagged information and the other to retrieve geotagged tweets. In our case, the queries were designed to maximize the number of tweets collected for each language. 
The first query searches for the most common words for one specific language, i.e., the so-called {\em stopwords}, that means, for instance, in English, we look for words like \emph{the, also, well, etc.}, in Spanish words like \emph{el, los, como, etc.} using both accented and unaccented version, if it applies. This way, we can download the maximum number of tweets for a specific language because we use common tokens around 400 language-dependent words allowed by Twitter queries. Meanwhile, the second query also includes a geo-query that includes the entire world's geographic coordinates, latitude and longitude values, and the language code.

\begin{figure}
\centering
\begin{subfigure}[b]{0.49\textwidth}
    \centering
    \includegraphics[width=1.0\textwidth]{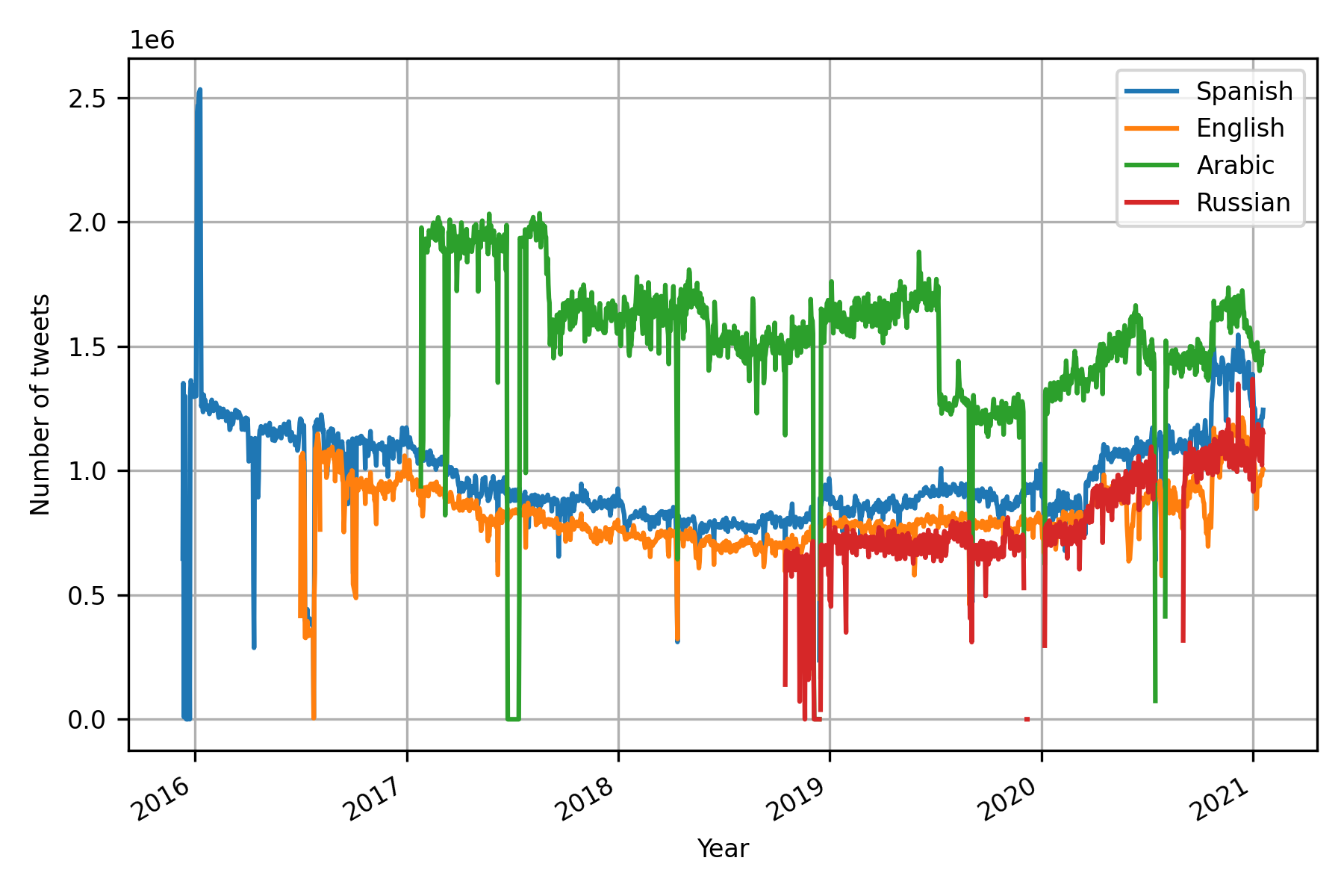}
    \caption{Tweets without geotagged information.}
    \label{fig:number_tweets}
\end{subfigure}
\begin{subfigure}[b]{0.49\textwidth}
    \centering
    \includegraphics[width=1.0\textwidth]{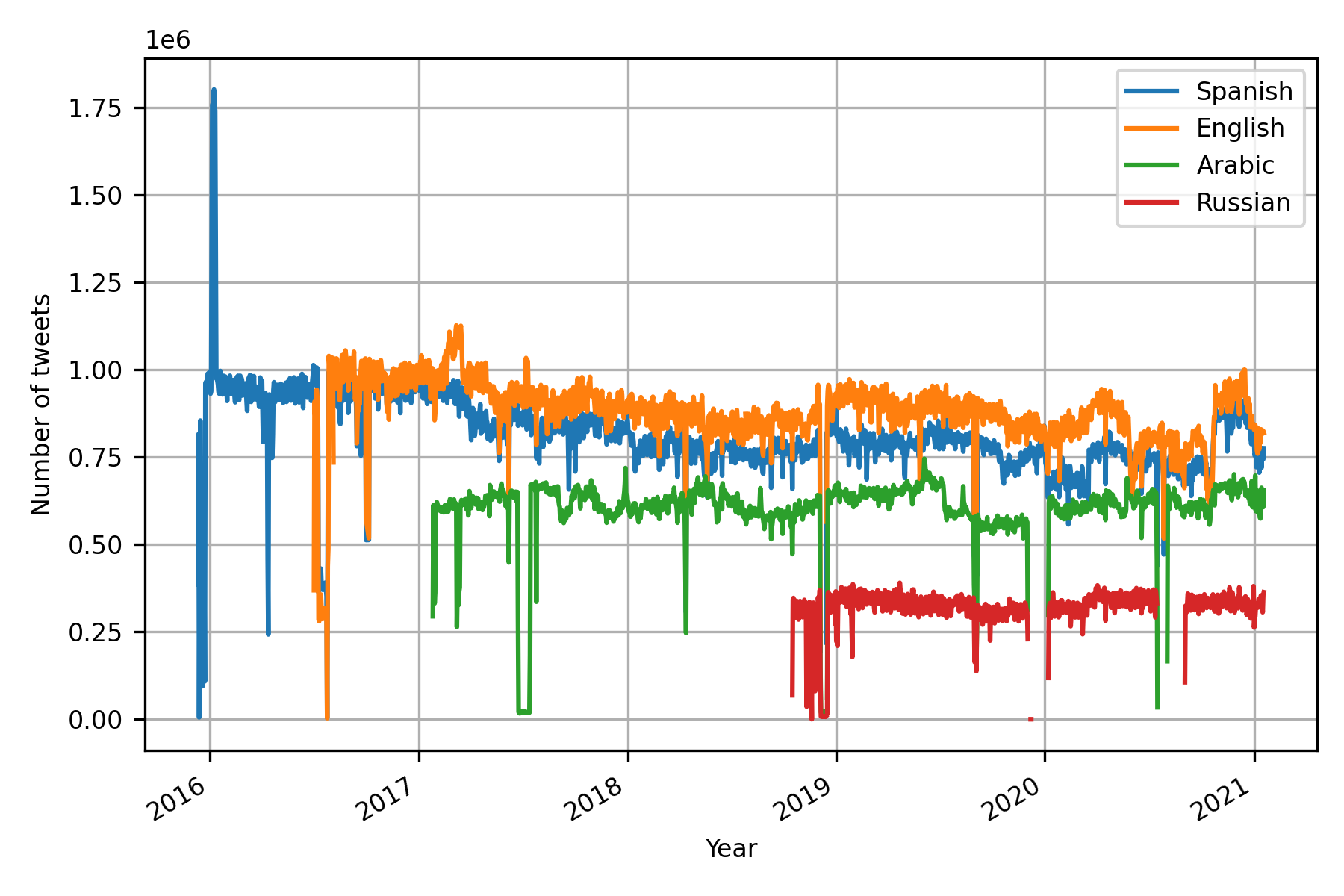}
    \caption{Geotagged tweets.}
    \label{fig:number_geo_tweets}
\end{subfigure}
\caption{Number of tweets for Spanish, English, Arabic, and Russian languages.}
\label{fig:tweets}
\end{figure}

Figure~\ref{fig:tweets} presents the number of tweets (in millions) that have been processed per day. Figure \ref{fig:number_tweets} presents the number of tweets retrieved without geotagged information; this was the first download process launched. Spanish was the first language being collected by us (December 12, 2015), followed by English (July 2, 2016), then Arabic (February 17, 2017), and the last one corresponds to the Russian language (October 31, 2018). From this figure, some particularities can be seen, such as the number of tweets for English and Spanish languages present a decreasing trend that stops around April 2017. This decreasing tendency may be caused by a substantial re-organization of the social media user distributions\footnote{The rise of social media, \url{https://ourworldindata.org/rise-of-social-media}; revised in July 2019.}. Please note that the decreasing slope is not present in the other languages since the collection started around or after April 2017. On the other hand, it can be observed that a bump arose in the English and Spanish languages around mid-February 2020, whenever the COVID-19 epidemic started in America. 

Figure~\ref{fig:number_geo_tweets} complements the information presenting the number of geotagged tweets (in millions) per day. As can be seen, the curves' characteristics are similar in both figures being the difference that the number of tweets with location data is less than the non-geotagged tweets. That means that most messages were not geolocalized by their authors. The geotagged tweets also provide the country information. To have an insight of the distribution, in 2019,  the majority of tweets had been produced in the United States (36\%), this is followed by Saudi Arabia (7\%), Great Britain (6\%), Argentine (6\%), Spain (5\%), Russia (5\%), Mexico (4\%), Egypt (2\%), Colombia (2\%), and Canada (2\%). The rest of the countries contribute to 25 \% of the total of geotagged tweets produced.

\subsection{Data's processing and cleaning}
\label{sec:cleaning}

The tweets collected are processed by day using the tokenizers of our previous development~\cite{Tellez2017ATwitter}. The process corresponds to splitting text into words and bi-grams, setting the text to lowercase, removing the punctuation symbols, numbers, and URLs. Blank symbols are replaced by symbol $\sim$, and, consequently, the words in a bi-gram are joined by it. Only those tokens that appear at least 0.01\% of the retrieved tweets per day are included. Finally, the tokens obtained for each language are grouped by country, if the country's information is present, and tweets without it. 

\subsection{Processing Mobility}
\label{sec:pmobility}

The geotagged information, as georeferenced tweets, can serve as a proxy to estimate people's mobility. The process used to estimate it can be divided into two stages. The first one was performed only once, and the second one is responsible for actual mobility values. The first step computes geographic landmarks that will provide the lowest resolution in the space. The idea is that each geotagged tweet will be associated with a landmark. The landmark set is a subset of the bounding box of all the tweets in the collection. The bounding box kept appeared in more tweets than 1\% of the number of days collected for the analyzed country. The idea of using the bounding box as landmarks and keeping only those used more frequently corresponds to protecting the positional information of users that submit their exact position to social media. Nonetheless, in estimating mobility, those messages with exact location are considered when the distance between two consecutive tweets is longer than 100 meters. The movement is assigned to their corresponding landmarks. 

\begin{figure*}
    {\includegraphics[trim=5.5cm 0 5.5cm 0, clip, width=1.0\textwidth]{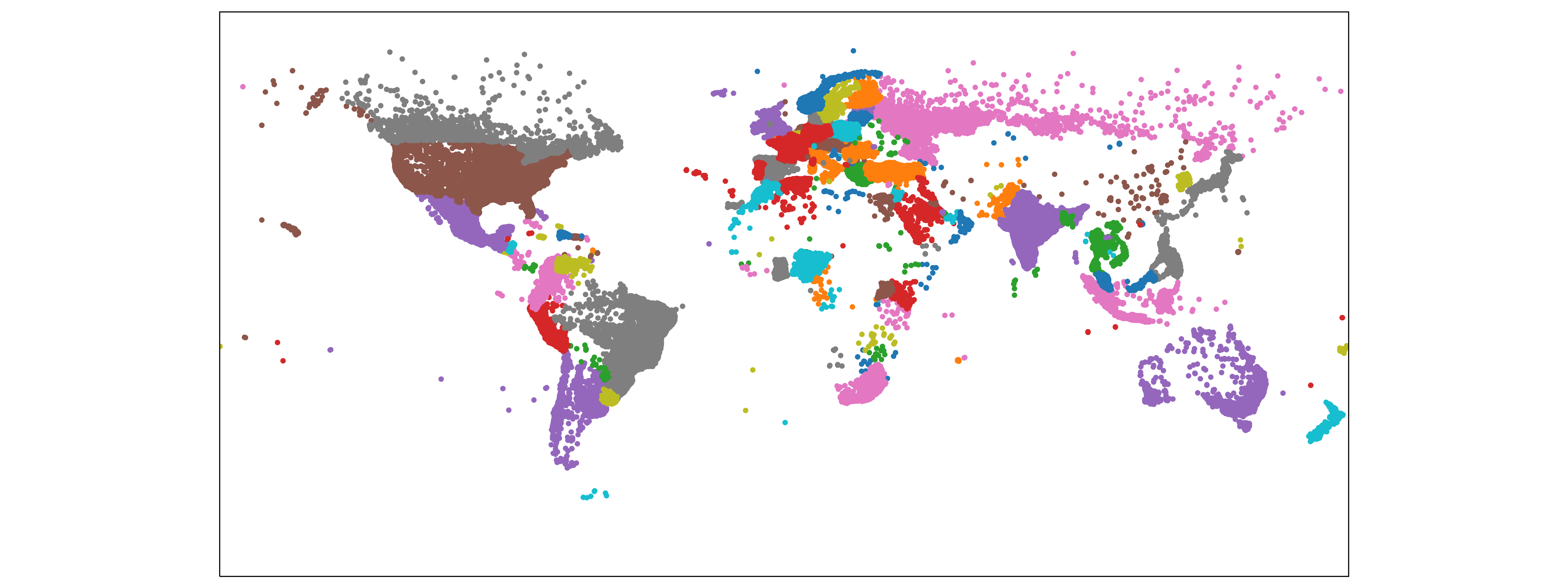}}
    \centering
    \caption{Set of geographic points used to calculate mobility.}

    \label{fig:tweet-map}
\end{figure*}

\begin{figure}
    \centering
    \includegraphics[width=0.49\textwidth]{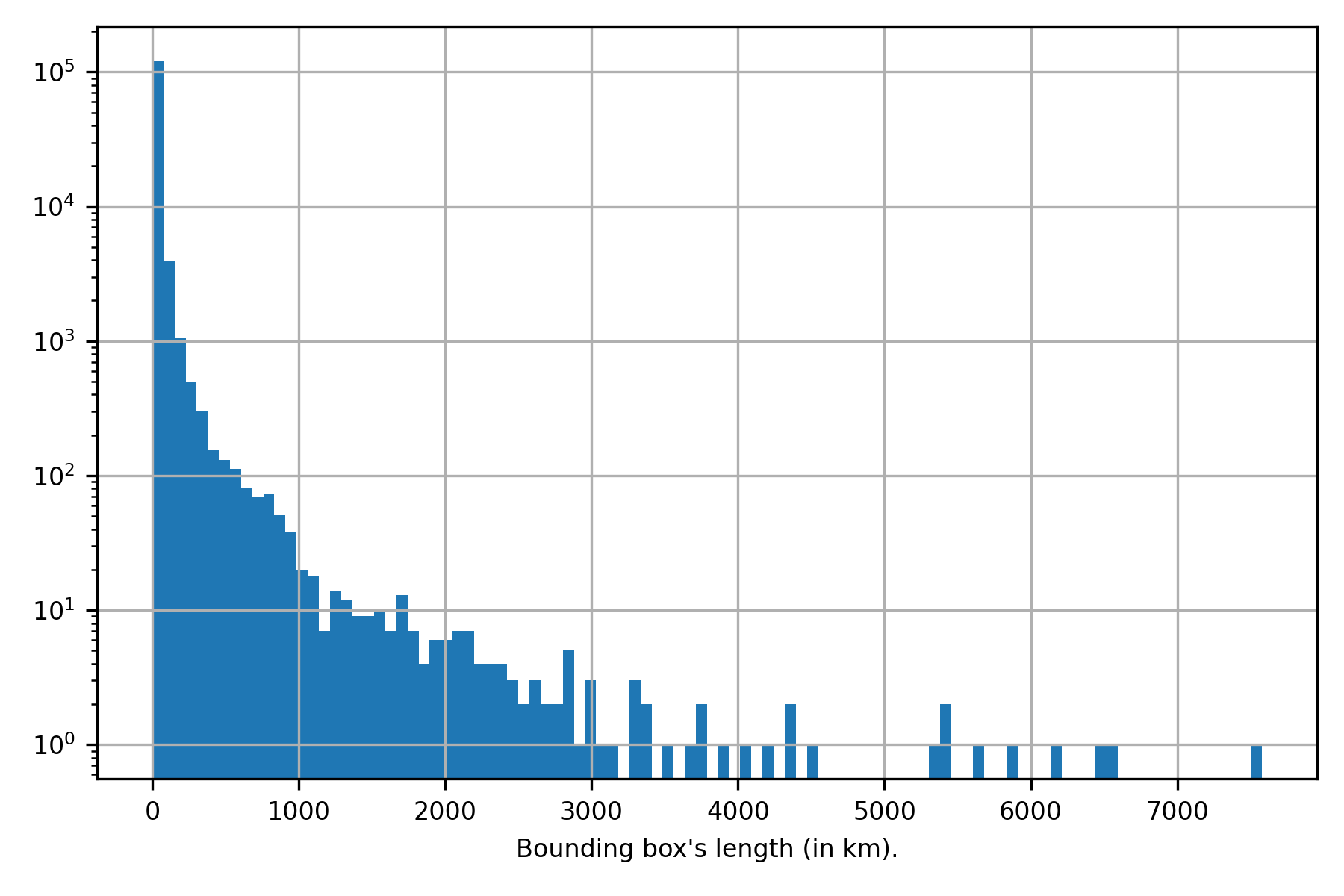}
    \caption{Histogram of the bounding box's length used as landmarks.}
    \label{fig:histogram-bounding-box-length}
\end{figure}

The landmark set, where each point corresponds to a bounding box's centroid, is depicted in Figure \ref{fig:tweet-map}, where all the points belonging to the same country have the same color; the result is that the countries can be observed in the picture. Complementing the information provided, Figure \ref{fig:histogram-bounding-box-length} presents a histogram of the bounding box's length; it can be observed that most bounding boxes are on the first bin, which corresponds to a length less than one kilometer.

The second step corresponds to the count of the trips between the elements of the landmark set, previously defined (first step). Each landmark is represented with a unique identifier. Then, each geotagged tweet is associated with a landmark. There are three cases to perform this association; the more frequently one is when the tweet's bounding box is in the landmark set. On the other hand, when the tweet information is a bounding box not in the set or a point location, it is associated with the landmark set's closest element; this latter operation is performed using the centroids of the bounding boxes. The next step grouped the tweets by the user and sorted them by time. The process iterates for all the tweets of a user; comparing the position of each tweet with the place of the following tweet; in the case, the length between the two locations is higher than 100 meters, then it is recorded as one trip from the first tweet identifier to the second tweet identifier. This process continues for all the users that published some message on that day. A limitation of this algorithm is that it does not consider trips that occur on different days, only on the same day. However, this decision was taken to favor efficiency in terms of time -- the algorithm can be run in parallel at the period of a day -- and memory since the number of users increases when one considers more days.

\begin{figure}
    \centering
    \includegraphics[width=0.49\textwidth]{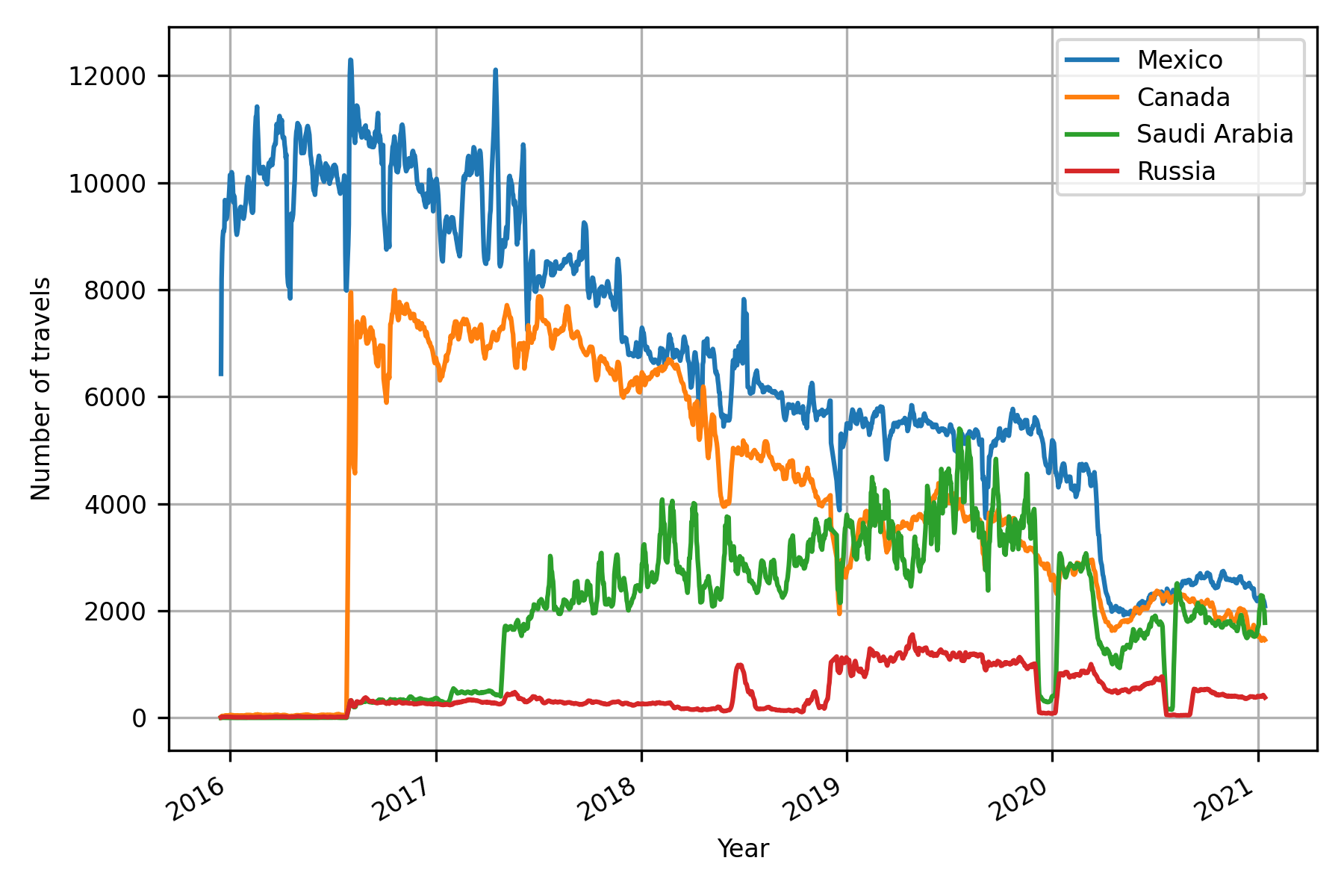}
    \caption{Mobility of Mexico, Canada, Saudi Arabia, and Russia.}
    \label{fig:mx-ca-sa-ru}
\end{figure}

The number of trips is the starting point to compute the mobility information that can be measured as the number of outward, inward, inside travels and the aggregation of all mobility events, namely overall mobility. Different measures can represent mobility; these could be, for example, the number of travels or the mobility trend seen as a percentage between the difference of a baseline and a specific date. Figure \ref{fig:mx-ca-sa-ru} presents the overall mobility (number of travels) of four countries which are representative of the different languages collected, namely Mexico (Spanish), Canada (English), Saudi Arabia (Arabic), and Russia (Russian). The idea is to depict the period that can be queried from the library; as seen from the figure, mobility starts for Spanish-speaking countries, followed by English-speaking countries.  It can be observed that the mobility of Saudi Arabia and Russia starts in the same period that Canada; this is because there are tweets in English on those countries that are used to estimate the mobility. On the other hand, there is an increase in Saudi Arabia's mobility when the Arabic language is retrieved from Twitter (see Figure \ref{fig:number_geo_tweets}); an equivalent behavior is seen for Russian.
There is another feature in Figure \ref{fig:mx-ca-sa-ru} that we would like to draw attention to, which is the effect that COVID-19 has had on the mobility pattern. It can be observed a decreasing trend in mobility that started around March 12, 2020. The lowest point appears one month later and then starts a positive slope, followed by a negative slope around late November 2020. These correspond to the first and second waves of the pandemic. 

Another mobility measure that has been widely used is to present mobility as the percentage of change, transforming the number of trips into a rate by considering a baseline period. A plausible approach followed in different research works is to compute the average of mobility per weekday. Usually, the baseline period is previous to the period of interest. For example, Facebook Disaster Maps \cite{Facebook2019FacebookRecovery} uses 5-13 weeks previous to the crisis to compute the baseline. The percentage is computed using the baseline information of the same weekday and hour period. We follow this approach to present mobility as a percentage. The value used is the median mobility per weekday; the median is obtained using 13 weeks previous to the period of interest.

Using a statistic per weekday to compute the percentage relies on the fact that the mobility pattern depends on the weekday. However, there are situations where this is not the case; for example, the mobility of a long weekend Monday is more similar to a Sunday than to a standard Monday. Instead of identifying each day's particular characteristics and then developing an algorithm that considers these differences, one can abandon the idea that mobility can be grouped by weekday and use mobility with a clustering algorithm to find the groups automatically. 

The idea of automatically finding the groups is explored using the well-known k-means algorithm. The mobility in the baseline period is used to train a k-means (where $k$ corresponds to the highest value of the Silhouette score obtained by varying $k \in [2, 7]$). The percentage of mobility is computed as the rate between the closest centroid and the mobility value at hand; this process facilitates the comparison between the two procedures to compute the percentage.

\section{Examples of the library's usage for analysis}
\label{sec:analysis}

The library's examples presented in this section are about two aspects, the first one, the text  (tweets' content), the second one, the mobility computed with the geographic information of tweets. These two types of data are included in the proposed library.

\subsection{Using text}
\label{sec:text}

\begin{figure*}
\centering
\begin{subfigure}[b]{0.45\textwidth}
    \centering
    \includegraphics[width=1.0\textwidth]{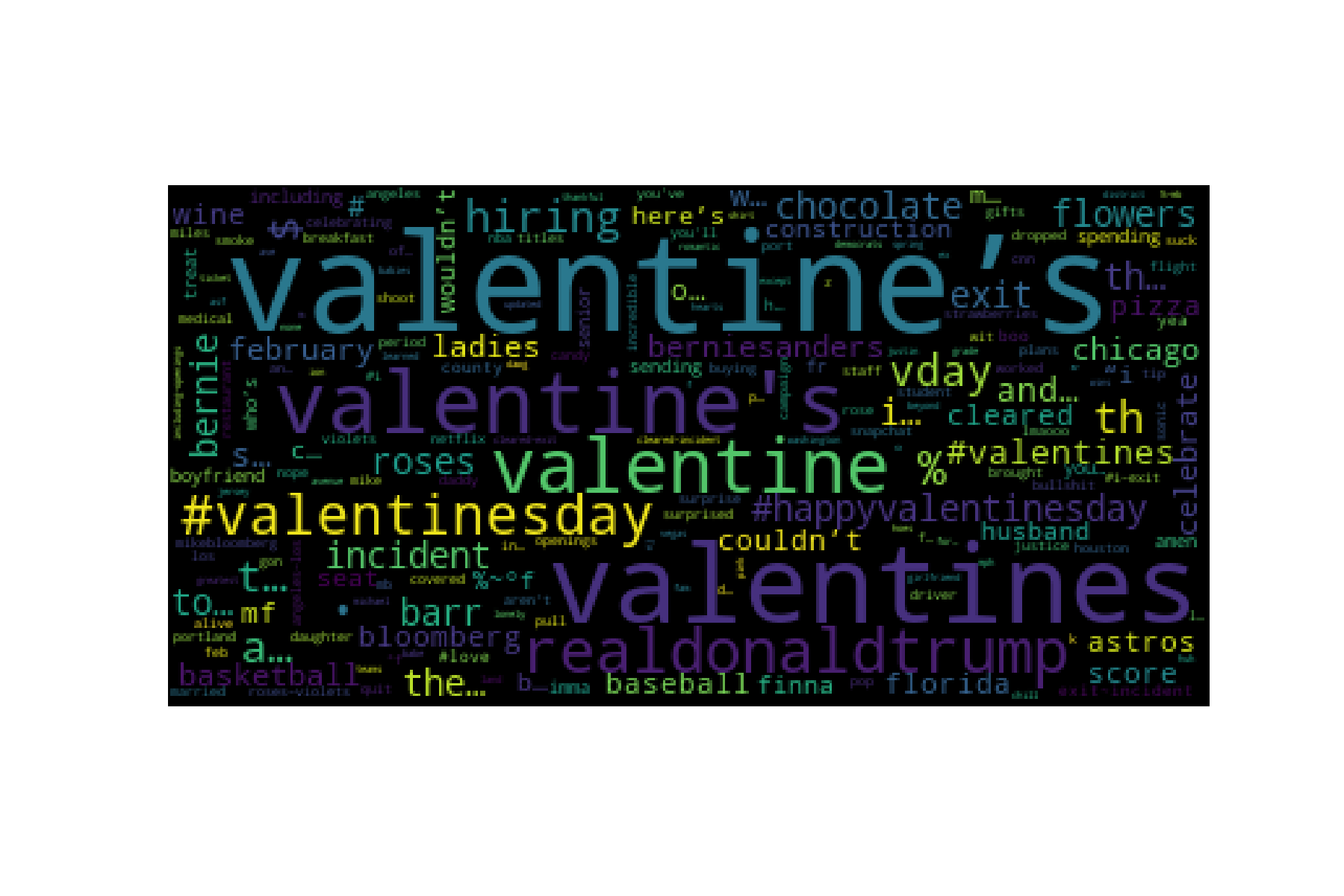}
    \caption{United States word cloud of February 14, 2020. Valentine's Day.}
    \label{fig:word-cloud-us}
\end{subfigure}\hspace{2em}\begin{subfigure}[b]{0.45\textwidth}
    \centering
    \includegraphics[width=1.0\textwidth]{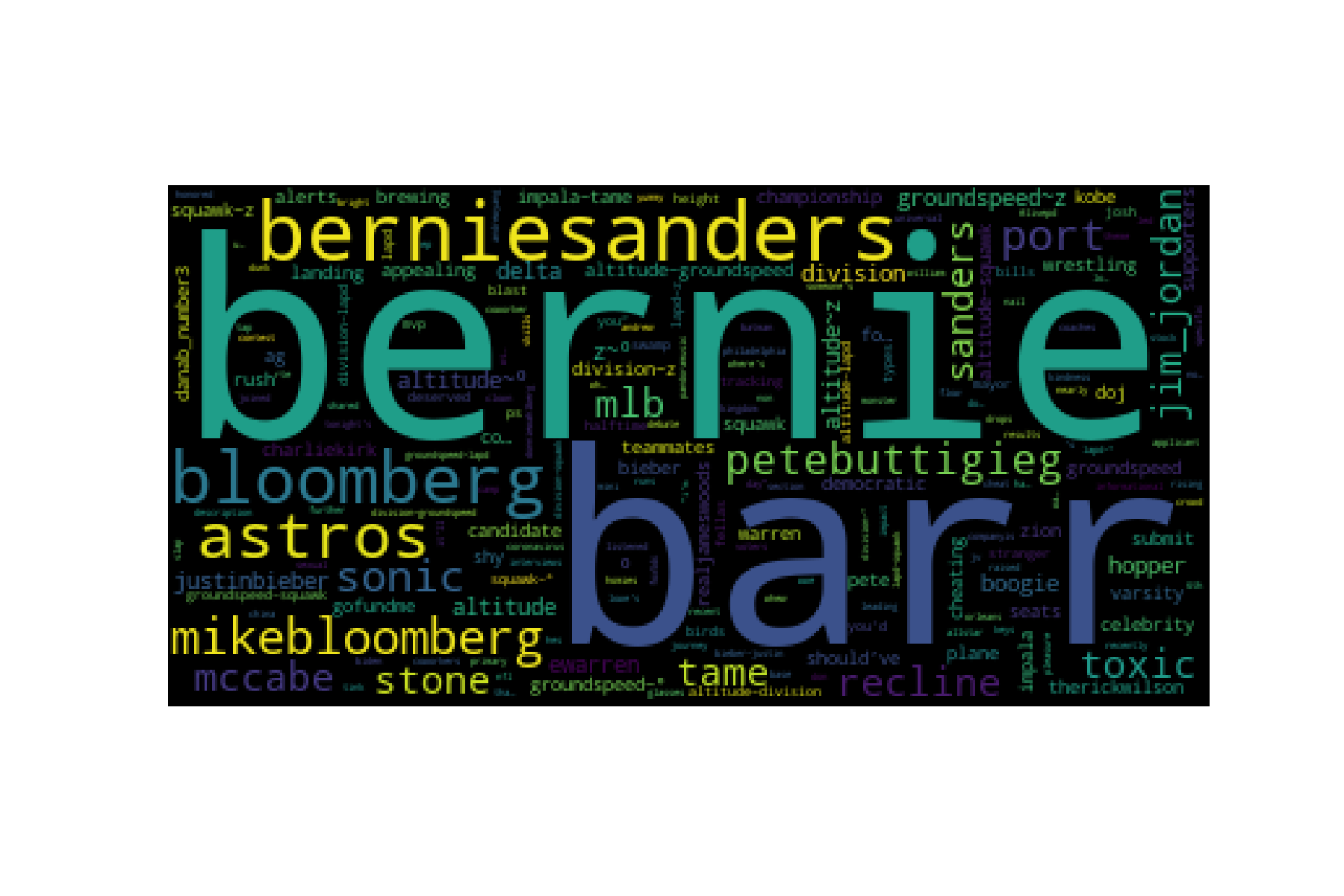}
    \caption{U.S. word cloud (2020-02-14) without tokens of previous years.}
    \label{fig:word-cloud-us-2}
\end{subfigure}
\caption{United States word clouds obtained by removing frequent tokens, and emojis.}
\label{fig:word-cloud}
\end{figure*}

Tweets are text messages limited to at most 280 symbols and contain the information that users share. Analyzing this information could give a huge insight into any event or social phenomenon. Even only seeing the word frequencies over time may be helpful. In this sense, the word clouds are a useful and well-known tool to visualize these word frequencies easily. Figure \ref{fig:word-cloud} presents word clouds for the United States on two different days. The word clouds were obtained by removing the frequent tokens\footnote{Frequent tokens are the vocabulary obtained from randomly selecting 5,000,000 tweets from the whole collection and keeping only those tokens that appeared at least 0.1\% of the time.}. Emojis were also removed due to compatibility issues on the plotting library used to create the word cloud. Figures \ref{fig:word-cloud-us} and \ref{fig:word-cloud-us-2} show the United States' word cloud on February 14, 2020.
%whereas Figures \ref{fig:word-cloud-mx} and \ref{fig:word-cloud-mx-2} depict Mexico's word cloud on May 10, 2020, which corresponds to the celebration of the Mother's Day in Mexico. 
We want to remark that the word cloud's functionality is included in \emph{text models} library, which, in turn, uses the \textit{wordcloud} library\footnote{https://pypi.org/project/wordcloud/}. Nevertheless, any other word cloud library can also be applied to the data generated with \emph{text models} library.

As can be seen in Figure \ref{fig:word-cloud-us} (left of the figure), word clouds represent only the holiday, and any other event is hidden behind it. That is, on Valentine's Day, the most common word is {\em valentine's} and its variations. A simple procedure to retrieve other events present on the holiday, and probably not related to it, is to remove the most frequent tokens (such as words) that appeared in the holiday's previous years. Removing the tokens of the previous years produce Figure \ref{fig:word-cloud-us-2}. It can be observed that the topic in the United States was about {\em Bernie Sanders}.

\begin{figure*}
\centering
\begin{subfigure}[b]{0.45\textwidth}
    \centering
    \includegraphics[width=0.9\textwidth]{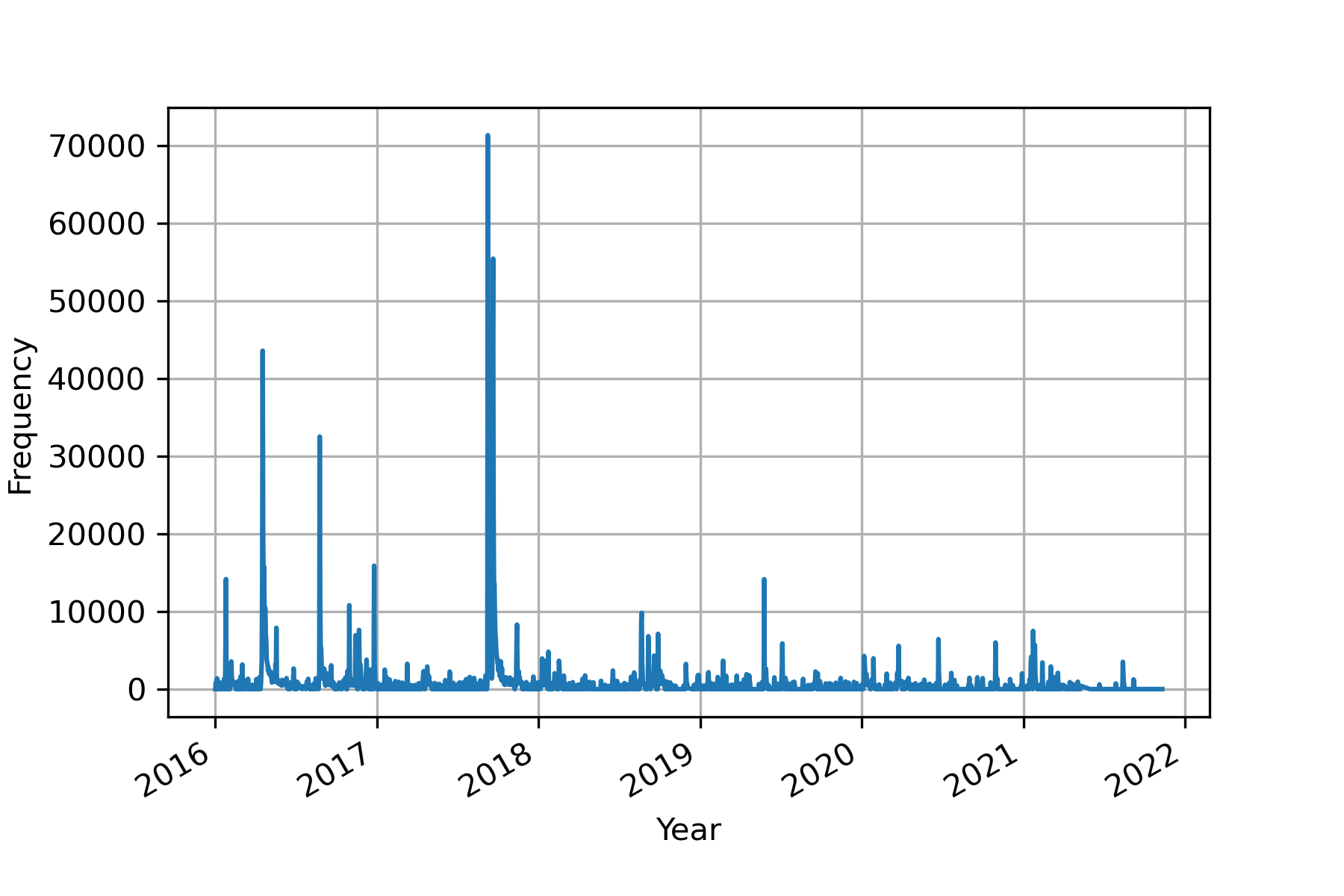}
    \caption{Frequency of {\em terremoto} from January 1st, 2016 to November 1st, 2021.}
    \label{fig:terremoto-freq}
\end{subfigure}\hspace{2em}\begin{subfigure}[b]{0.45\textwidth}
    \centering
    \includegraphics[width=1.0\textwidth]{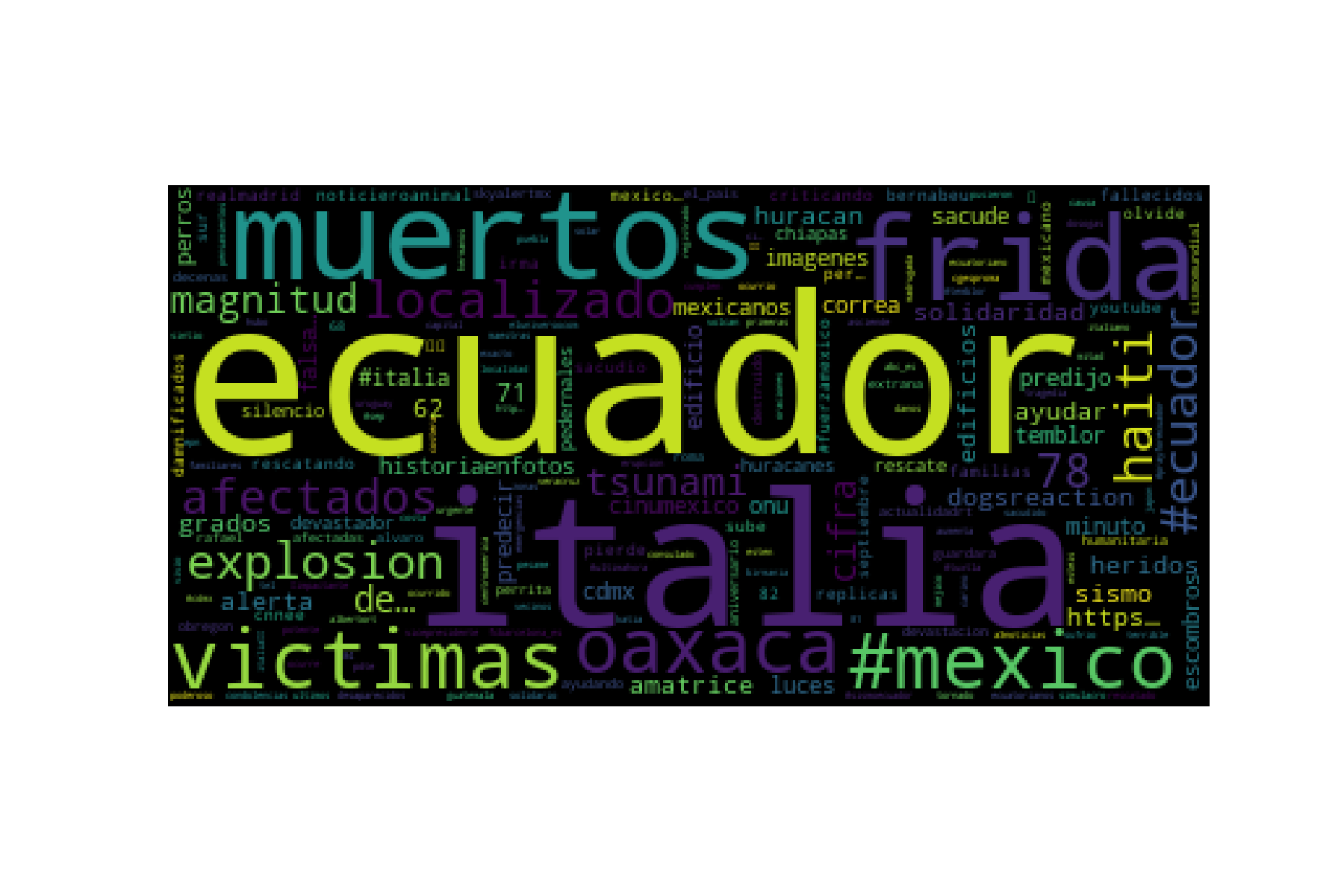}
    \caption{Word cloud of the {\em terremoto} co-current words for the days with a {\em terremoto}'s frequency higher than 20,000.}
    \label{fig:terremoto-wordcloud}
\end{subfigure}
\caption{Frequency of the word {\em terremoto} (earthquake) and word cloud of the co-current words.}
\label{fig:terremoto}
\end{figure*}

The previous approach depicts the case where one is interested in a particular date and wants to know the significant event removing a recurrent event such as Valentine's Day. The complement could be to know the dates where a particular event occurred. Let us illustrate this idea by retrieving the frequency of the word {\em terremoto} (earthquake) from January 1st, 2016 to November 1st, 2021. Figure \ref{fig:terremoto-freq} shows the frequency, and it can be observed that there are five days where the frequency is higher than 20,000. These dates correspond to April 17 and 18, 2016, August 8th, 2016, and September 8, 19, 20, and 21, 2017. We can examine more about these dates by looking at the co-occurrence of the vocabulary (excepting common words) with {\em terremoto}. Figure \ref{fig:terremoto-wordcloud} shows the word cloud of the co-occurrent words removing {\em terremoto}. The weights are normalized for the day, and for those words appearing in more than one day, the higher normalized frequency is kept. The words {\em ecuador} (Ecuador), {\em italia} (Italy), and {\em mexico} (Mexico) are present on the word cloud, which correspond to the three countries where the earthquake took place. The earthquake in Ecuador was in April 2016, in Italy was in August 2016, and there were two in Mexico, the first one September 8th and the second on September 19th, 2017. 

\subsubsection{Topic Modeling}
\begin{figure*}
\centering
\begin{subfigure}[b]{0.45\textwidth}
    \centering
    \includegraphics[width=1.0\textwidth]{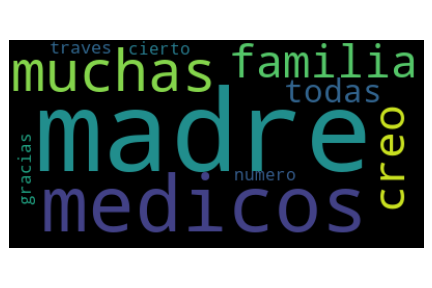}
    \caption{Most representative words on topic 1.}
    \label{fig:topics1}
\end{subfigure}\hspace{2em}\begin{subfigure}[b]{0.45\textwidth}
    \centering
    \includegraphics[width=1.0\textwidth]{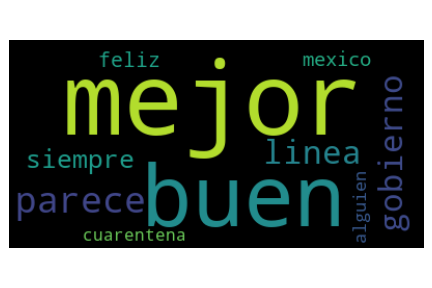}
    \caption{Most representative words on topic 2.}
    \label{fig:topics2}
\end{subfigure}
\begin{subfigure}[b]{0.45\textwidth}
    \centering
    \includegraphics[width=1.0\textwidth]{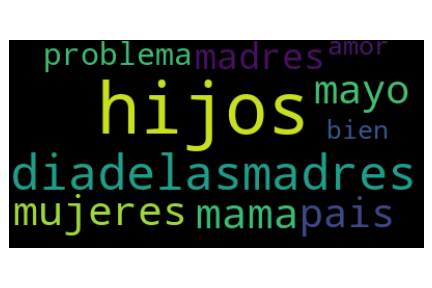}
    \caption{Most representative words on topic 3.}
    \label{fig:topics3}
\end{subfigure}\hspace{2em}\begin{subfigure}[b]{0.45\textwidth}
    \centering
    \includegraphics[width=1.0\textwidth]{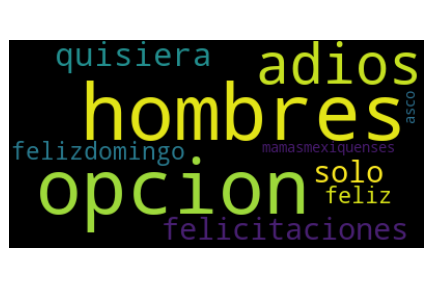}
    \caption{Most representative words on topic 4.}
    \label{fig:topics4}
\end{subfigure}
\caption{Word clouds of most representative 10 words in each generated topic on the date 2020-05-10 (Mother's Day) from Mexico.}
\label{fig:topics}
\end{figure*}

%LDA 
Another example of a well-known and widespread technique is \textit{topic modeling}.
To know what topics were included in a set of tweets, we use LDA (Latent Dirichlet Allocation) method~\cite{Blei2003LatentJordan}. LDA is a generative probabilistic model for data collections, such as tweets. LDA's idea is to group the text into topics, these topics have a representative set of words, and each document is then assumed to contain multiple overlapping topics. Each topic can be described by a set of words that hold the highest probability of association with that topic. We use Gensim\footnote{https://pypi.org/project/gensim/} library to compute the LDA model.
In this sense, the \emph{text models} library provides a vocabulary of terms with their frequencies. From that data, a corpus was generated to be the input of the LDA algorithm. This corpus is only the text repeated in its frequency indicated by \emph{text models} removing emojis. Here we show how to create the topic analysis illustrated in Figure~\ref{fig:topics}. The high-level steps for this exercise are as follows:
\begin{enumerate}
    \item A dataset using \emph{text models} library was extracted. In this case, we used tweets from Mexico on the date of 2020-05-10.
    \item Then, the emojis, and most common words were removed, also using the functions of the \emph{text models} library.
    \item Since \emph{text models} gives data as vocabulary, that means (word-frequency), that was used to generate the LDA algorithm's input.
    \item Pre-processing steps were done deleting links, white spaces, stop-words, and words equal or smaller than three characters. 
    \item Then an LDA analysis with $t$ topics was generated. For instance, we set $t=4$; that is, four topics were generated.
\end{enumerate}

As a result of topic modeling, four topics were generated for the 2020-05-10 date. This date corresponds to Mother's Day in Mexico; please note that the data is limited to the Mexican territory and Spanish language. Figure~\ref{fig:topics} shows the word clouds related to each one of these four topics. The figure shows words that could be related to Mother's Day; 
nevertheless, there are some words such as 
``m\'edicos'' and ``gobierno'' (doctors and government in English) are odd on this particular day, but it is perfectly understandable by the pandemic circumstances.

\begin{figure*}
    \centering
    \includegraphics[width=0.95\textwidth]{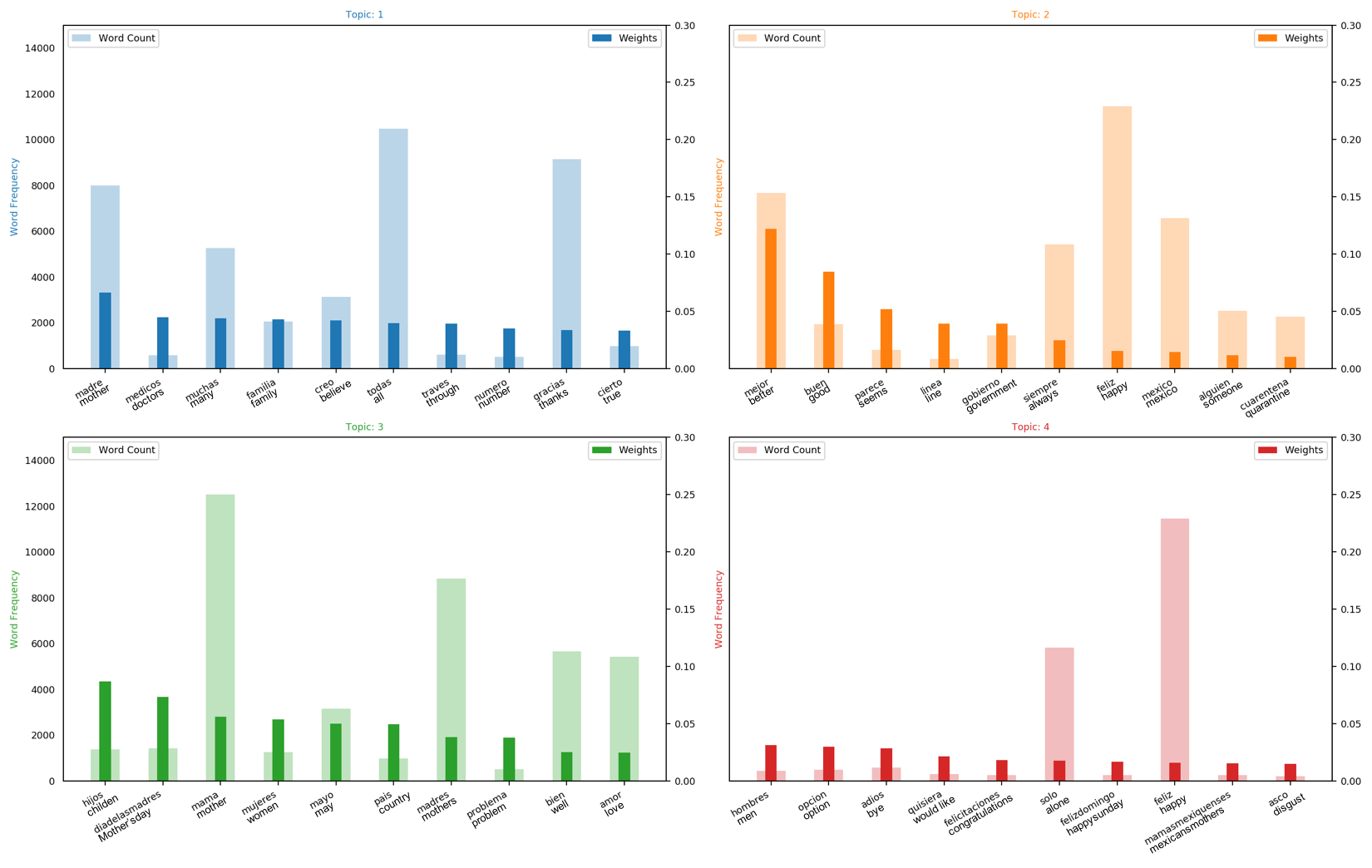}
    \caption{Keywords frequencies and their weights for each topic on date 2020-05-10 (Mother's Day) for Mexico's Spanish messages.}
    \label{fig:histogram_topics}
\end{figure*}

The topics can also be visualized using a histogram. These figures were done with Matplotlib package\footnote{see for more details https://matplotlib.org/}. Figure~\ref{fig:histogram_topics} shows words and their corresponding weight and frequency for each topic. Please note that high-frequencies do not necessarily imply large weights for the LDA model. The topic analysis allows us to discover patterns or events on Twitter.

\subsubsection{Similarity between Spanish-speaking Countries}

The different branches of a language are named dialects. In each of them, diverse terms could be used for other things or concepts. The study of these dialects can help us understand cultural aspects among regions and their closeness~\cite{Donoso2017DialectometricTwitter}. For instance, Twitter posts can serve as a proxy to study the similarity between Spanish language variations or languages.

\begin{figure}
    \centering
    \includegraphics[width=0.50\textwidth]{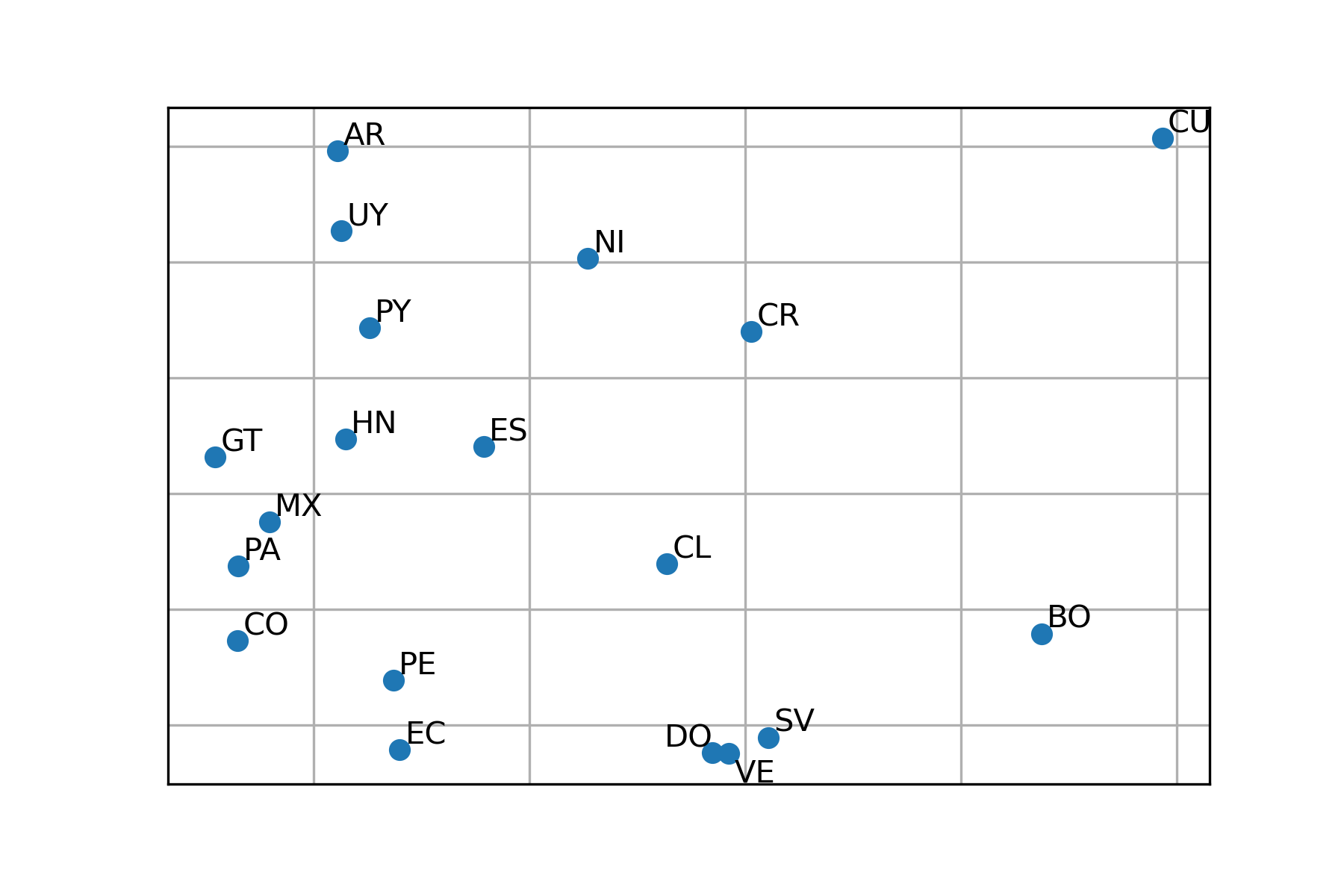}
    \caption{PCA's representation of Spanish-speaking countries computed on a matrix of the Jaccard similarity between words and bi-grams of words among the countries.}
\label{fig:similarity}
\end{figure}

Using the \emph{text models} library, the tokens and their frequency, grouped by country, can be used to model, for example, the similarity of a particular language in different countries. Figure \ref{fig:similarity} depicts the likeness of Spanish-speaking countries\footnote{Equatorial Guinea (GQ) is a Spanish-speaking country with other two official languages French and Portuguese, we did not include it in the study because there is not enough information}, i.e.,  Mexico (MX), Chile (CL), Spain (ES), Argentina (AR), Colombia (CO), Peru (PE), Venezuela (VE), Dominican Republic (DO), Paraguay (PY), Ecuador (EC), Uruguay (UY), Costa Rica (CR), El Salvador (SV), Panama (PA), Guatemala (GT), Honduras (HN), Nicaragua (NI),  Bolivia (BO), and Cuba (CU). 
The procedure to model language variations and compute their similarity was to create a vocabulary with words and bi-grams (of words) of 180 days selected randomly from January 1st, 2019, to November 1st, 2021. The vocabulary was created for each Spanish-speaking country. Once the vocabulary is obtained, the Jaccard similarity is used to measure all the pairs' similarities. The preceding process creates a similarity matrix transformed with Principal Components Analysis (PCA) to depict each country in a plane. 

Figure \ref{fig:similarity} depicts all Spanish-speaking countries; it can be observed that the most different country is Cuba (CU), which is also one of the countries where less information is retrieved from Twitter. From the figure, it can be observed that Mexico (MX) is closest to Panama (PA), 
Honduras (HN) and Guatemala (GT). Argentina (AR) and Uruguay (UY) can be seen as a cluster; these countries share a border and are located in the southern of South America. On the other hand, there is another cluster formed by Peru (PE), and Ecuador (EC) close to these is Colombia (CO). Another cluster is composed of El Salvador (SV), Venezuela (VE), and Dominican Republic (DO). This last cluster is not geographically related. The rest of the countries cannot be seen as belonging to a cluster. This experiment confirms that geographic boundaries are important when speaking of language similarities. This and other exciting questions could be answered using the \emph{text models} library and the provided data.  

\subsection{Mobility analysis}
\label{sec:mobility}

\begin{figure}
\centering
\begin{subfigure}[b]{0.49\textwidth}
    \centering
    \includegraphics[width=1.0\textwidth]{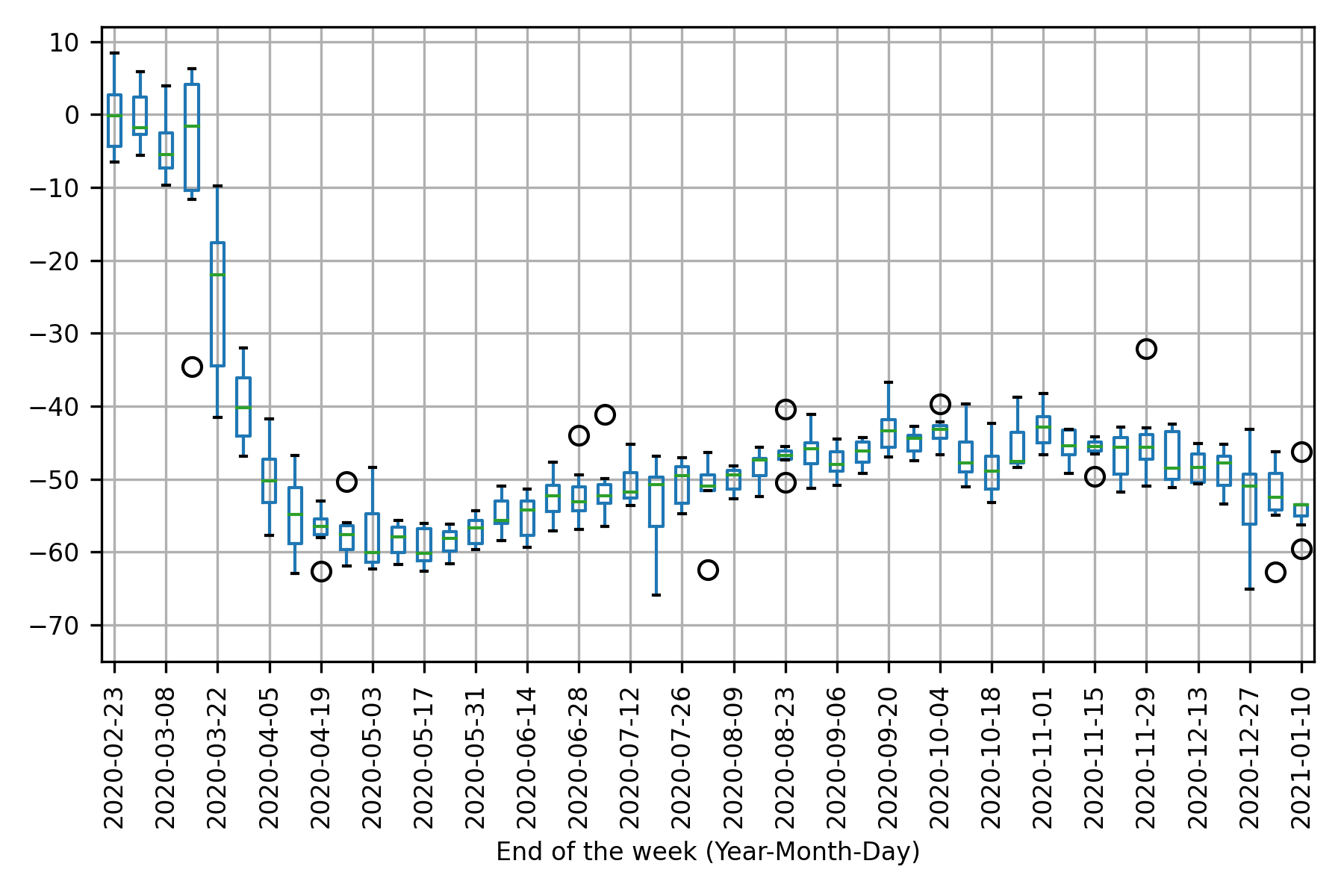}
    \caption{Computing percentage with weekday.}
    \label{fig:mexico_mobility_weekday}
\end{subfigure}
\begin{subfigure}[b]{0.49\textwidth}
    \centering
    \includegraphics[width=1.0\textwidth]{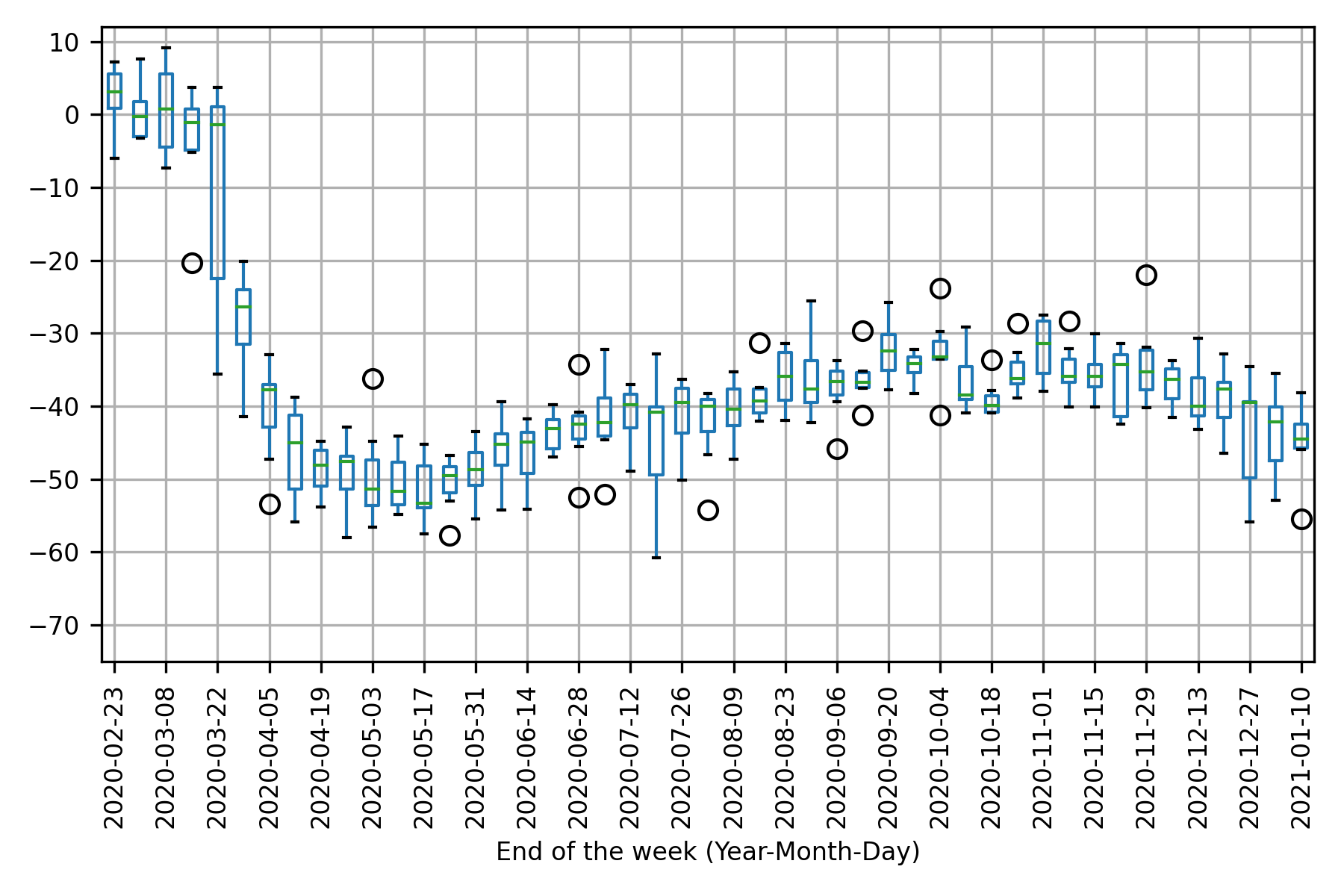}
    \caption{Computing percentage with k-means.}
    \label{fig:mexico_mobility_cluster}
\end{subfigure}
\caption{Boxplot of the Mexico's mobility on percentage.}
\label{fig:mexico_boxplot}
\end{figure}

The mobility examples start presenting a comparison between computing the percentage using the weekdays as a baseline and the k-means algorithm. Figure \ref{fig:mexico_boxplot} presents boxplots of Mexico's mobility, where each boxplot is obtained with the mobility of a week, and the date is the last day of the week.  

Figure \ref{fig:mexico_mobility_weekday} presents Mexico's mobility computing the percentage with the weekdays. Before March 22, 2020, it can be observed that the median is below -10\%, and there are weeks where the minimum value is almost -40\%. On the other hand, Figure \ref{fig:mexico_mobility_cluster} shows mobility when the percentage is computed with the centroids obtained with k-means. In the latter case, both first and third quantiles are close to zero percent before March 22, 2020. There is an outlier on the week ending on March 15, 2020, close to -20\%. After March 22, 2020, the lowest valley is below -60\% on Figure \ref{fig:mexico_mobility_weekday}, and the median is above -60\% on Figure \ref{fig:mexico_mobility_cluster}. As can be seen, there is a difference of approximately 10\% in the two procedures. This remarks on the importance of calculating and defining the baseline pattern of interest to observe the changes.

Typically, it would not be possible to compare the mobility obtained from one social media with another; however, given the outbreak of COVID-19, Google has generated a mobility report~\footnote{https://www.google.com/covid19/mobility/} that can be used for this purpose. Google's mobility used starts on February 12, 2020, and ends on January 1, 2021. The procedure used is to estimate the mobility with {\em text models} library for the same period as Google's mobility report. The mobility of Spanish-speaking and English-speaking countries is compared using Pearson correlation. The mobility values are smooth using a moving average of 7 days previous to computing the statistic.  

\begin{figure}
\centering
\begin{subfigure}[b]{0.49\textwidth}
    \centering
    \includegraphics[width=1.0\textwidth]{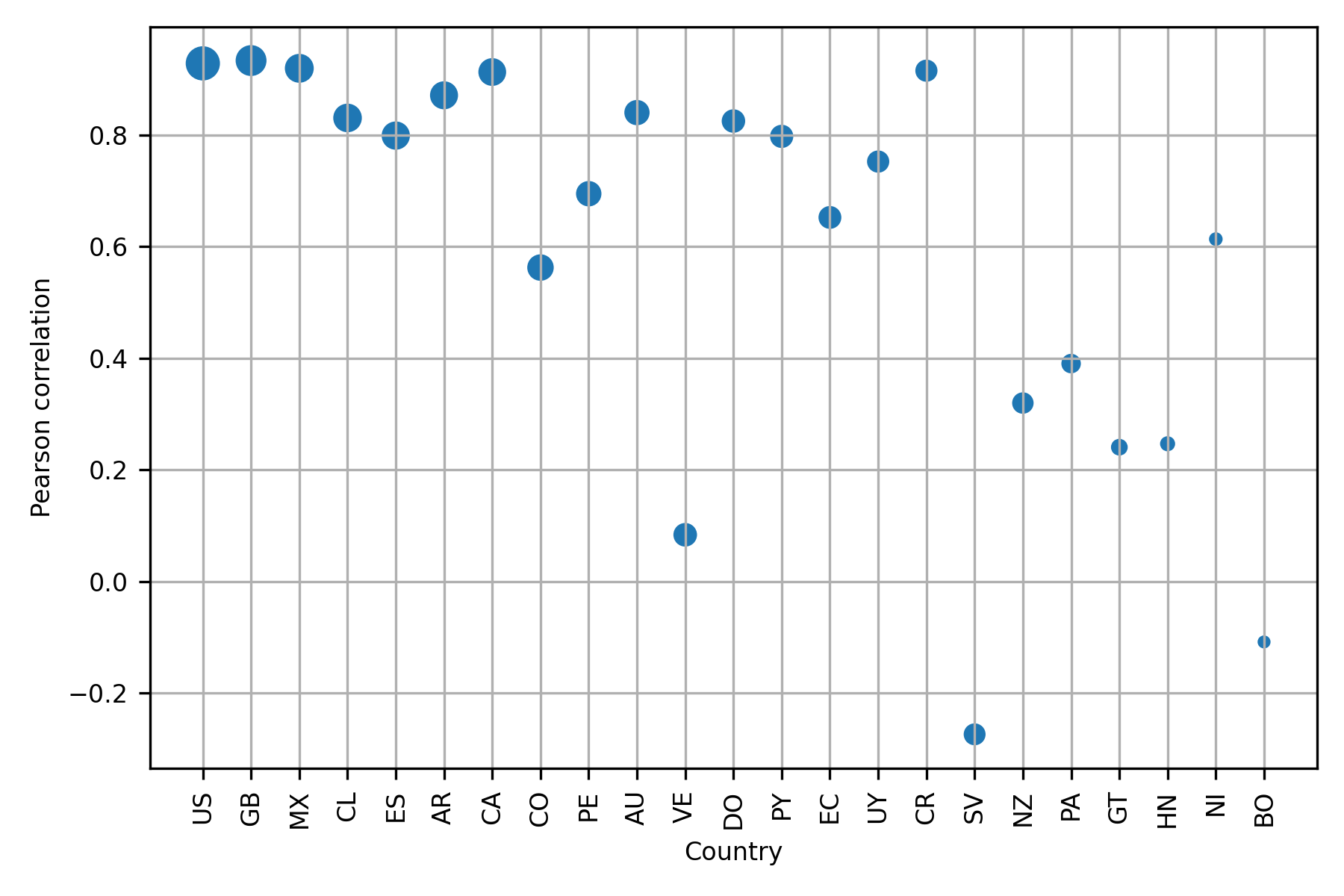}
    \caption{Mobility percentage using weekday.}
    \label{fig:gtc-weekday}
\end{subfigure}
\begin{subfigure}[b]{0.49\textwidth}
    \centering
    \includegraphics[width=1.0\textwidth]{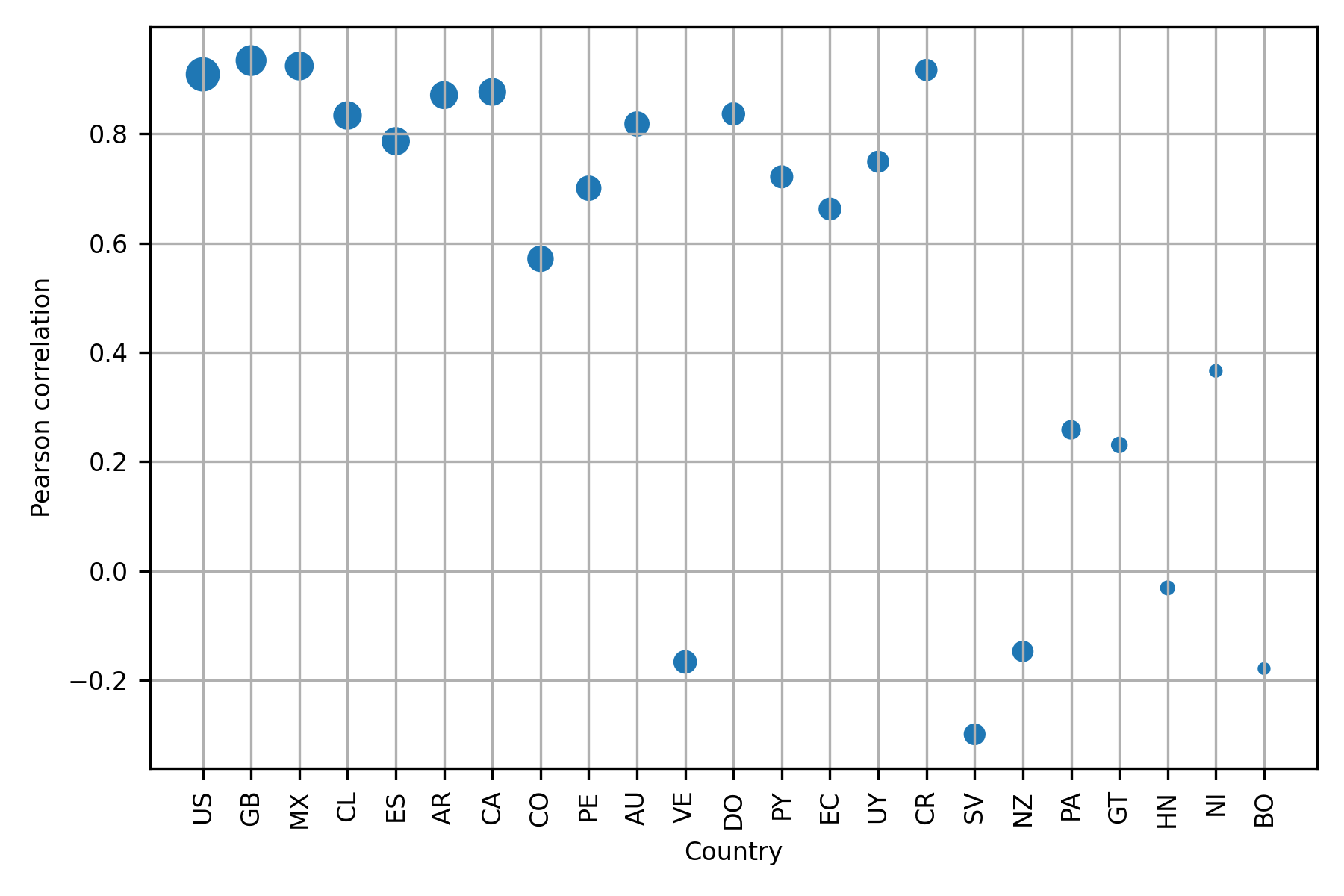}
    \caption{Mobility percentage using k-means.}
    \label{fig:gtc-kmeans}
\end{subfigure}
\caption{Correlation between Google's mobility report and the one calculated using Twitter data on different countries.}
\label{fig:google-twitter-correlation}
\end{figure}

Figure \ref{fig:google-twitter-correlation} presents the result of this correlation analysis on the United States (US), Great Britain (GB), Mexico (MX), Chile (CL), Spain (ES), Argentina (AR), Canada (CA), Colombia (CO), Peru (PE), Australia (AU), Venezuela (VE), Dominican Republic (DO), Paraguay (PY), Ecuador (EC), Uruguay (UY), Costa Rica (CR), El Salvador (SV), New Zealand (NZ), Panama (PA), Guatemala (GT), Honduras (HN), Nicaragua (NI), and Bolivia (BO). These countries correspond to the Spanish-speaking countries and English-speaking countries.  Cuba and Equatorial Guinea are not part of the comparison due to the lack of enough data. The markers' size indicates using a logarithm scale, the median of the number of travel in the baseline, and this number of countries.

The mobility obtained from Twitter corresponds to the percentage of change computed with weekdays and the percentage computed with k-means. From the figures, countries that present a correlation above $0.8$, and the lowest correlations correspond to the countries with fewer travels, taking into consideration the number of travels. Nonetheless, Venezuela is the exception with a median of 473 travels, whereas El Salvador (SV) has a median of 248 travels. From El Salvador to Bolivia (with a median of 22 travels), the correlation is below 0.8. Comparing the results presented in Figure \ref{fig:gtc-weekday} and \ref{fig:gtc-kmeans}, it is observed that in the majority of cases, the correlation obtained using the median of the weekday is higher than using the k-means algorithm. 

\begin{figure*}
    \centering
    {\includegraphics[trim=7cm 0 4cm 0, clip, width=1\textwidth]{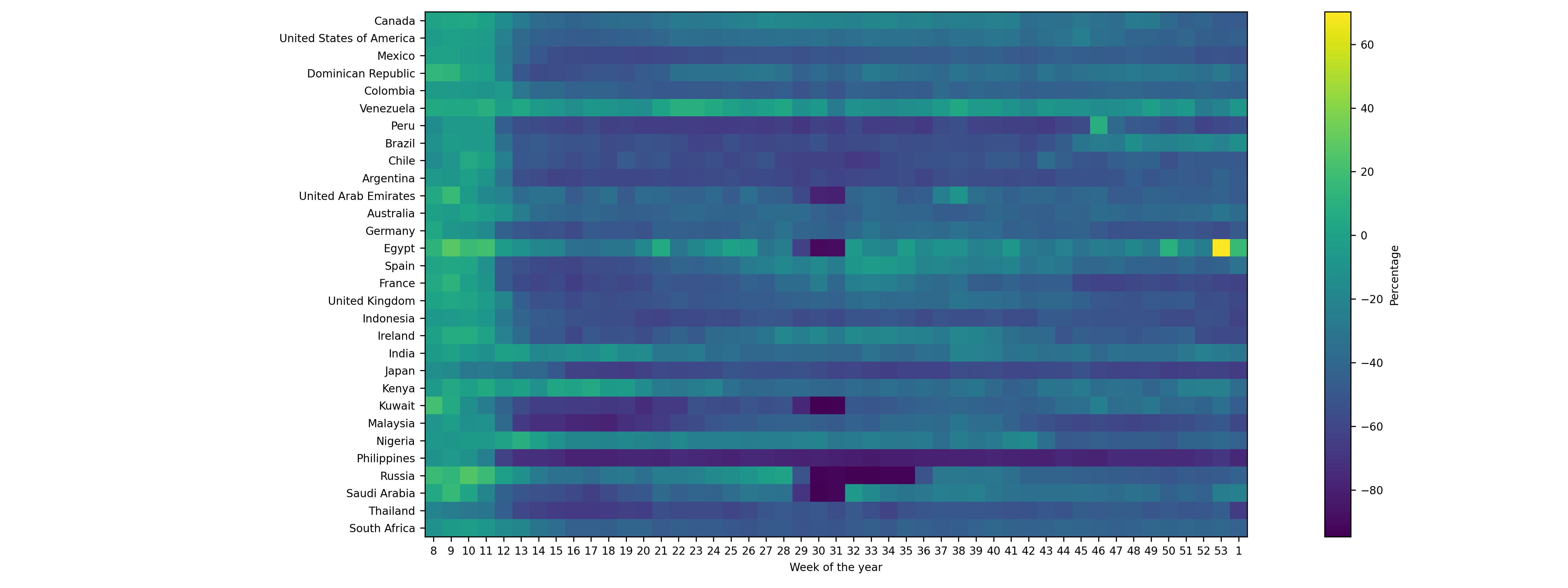}}
    \caption{Heatmap of the average mobility of a week of the 30 countries with more travels starting from week eight of 2020 until the first week of 2021.}
    \label{fig:world-mobility}
\end{figure*}

A plausible approach to visualize mobility in different countries is creating a heat map where the color is associated with mobility. The rows correspond to the time, and each column represents a country. Figure \ref{fig:world-mobility} presents this straightforward approach by computing the average mobility (as a percentage) of a week on the 30 countries with more travels. The countries are organized geographically, starting from the north if the country is in America, and the rest are set alphabetically.  The reduction of mobility on weeks 30 and 31 in Arabic and Russian-speaking countries is related to a problem with the tweets' collection (lack of data). From the figure, it can be observed that there were restrictions in mobility in most countries starting from week 12, and the restrictions continued for several weeks until these restrictions were released. Nonetheless, in the reported period, the countries did not return to the mobility presented on the week before week 12.

\section{Conclusions}
\label{sec:conclusions}

We presented the \emph{text models} library, which retrieves a plethora of information collected from Twitter in terms of tokens and their frequencies, and the mobility measured in different countries. The information is collected using Twitter's public API, where the query tries to maximize the number of tweets obtained in Arabic, English, Spanish, and Russian languages. 

The library aims to facilitate a newcomer, the data mining of Twitter data serving as an exploratory data analysis tool to know whether a particular event is reflected on Twitter or whatever hypothesis the user wants to demonstrate or explore. The capabilities of the library are illustrated with different applications performed with the library. We started showing how to create word clouds in different languages, from different countries, and on different dates. Additionally, we show how to use the information from previous years can highlight events that, in other circumstances, are opaque by a more popular one. As another example, we show used text messages to describe and contextualize an event via topic modeling. On the other hand, we used data retrieved from the library to analyze the difference between different Spanish language variations; for instance, we compare countries with a sizeable Spanish-speaking basis. Our experiment shows a correlation between language and geographic neighboring.

We show how to produce a mobility report on Mexico, Canada, and Saudi Arabia. The mobility is presented as percentages, and two procedures were described to perform this transformation. The first one corresponds to weekday statistics and has been used in the literature; however, using a clustering algorithm complements this idea tackling an issue seen at the beginning of the period analyzed. We also compared the mobility report obtained against Google's mobility report; the results show that there is a strong correlation between them; however, some countries are uncorrelated.

As is known, some limitations about the quality of Twitter data could be affected in any analysis or insights acquired with it. Nevertheless, this data has turned into valuable input in many research areas. So, the users of the proposed library and Twitter data, in consequence, must be aware of that.

A research avenue we would like to explore in future work is incorporating semantic representations in the library. The information provided can be used to create semantic models that can compare variations in the language concerning the country and time. 

In order to fulfill the aim of this contribution, in the appendix, we describe the code needed to replicate some of the applications, in particular, Figures \ref{fig:word-cloud-us}, \ref{fig:word-cloud-us-2}, \ref{fig:similarity}, \ref{fig:mx-ca-sa-ru}, and \ref{fig:mexico_mobility_weekday}. We consider that for a person familiar with Python, the work needed to replicate them is acceptable, and in most cases, a few code lines are enough to produce the desired output.

\appendix
\section{Library usage}
\label{sec:appendix}

The appendix aims to illustrate the use of the \emph{text models} library by replicating some of the applications presented in this manuscript. Specifically, Figures \ref{fig:word-cloud-us}, \ref{fig:word-cloud-us-2}, \ref{fig:similarity}, \ref{fig:mx-ca-sa-ru}, and \ref{fig:mexico_mobility_weekday} are replicated using the library which correspond to the applications presented using the tokens and their frequency and the mobility information. The first step is to install the library; the mobility examples follow this, and the last part corresponds to the use of tokens.

\subsection*{Preliminaries}

\emph{Text models} library can be installed by different means; however, the easiest way is to use the pip package installer. The following code installs the library using the command line.

\begin{lstlisting}[language=Bash]
pip install text_models
\end{lstlisting}

\subsection*{Mobility}

Once the \emph{text models} library is installed in the system, it can be used to retrieve the mobility and the vocabulary used at different periods of time. To illustrate the library's use, let us replicate Figure \ref{fig:mx-ca-sa-ru}, where the mobility is presented on the period contemplating from December 16th, 2015, to January 13th, 2021. The following code retrieved the mobility information on the specified period. 

{\small
\begin{lstlisting}[language=Python]
from text_models import Mobility
end=dict(year=2015, month=12, day=16)
start=dict(year=2021, month=1, day=13)
mob = Mobility(start, end=end)
\end{lstlisting}
}

Figure \ref{fig:mx-ca-sa-ru} presents mobility as the number of travels in Mexico, Canada, Saudi Arabia, and Russia. The following code lines compute mobility in all the countries. The first line counts the trips that occurred within the country and the inward and outward movement. The information is arranged in a {\em pandas' DataFrame} if the {\em pandas}' flag is activated; if the flag is deactivated, then dictionary objects are used. The second line retrieves the mobility for the countries of interest, i.e., Mexico (MX), Canada (CA), Saudi Arabia (SA), and Russia (RU). The third line plots the mobility and performs a moving average of seven days. 

{\small
\begin{lstlisting}[language=Python]
data = mob.overall(pandas=True)
m = data.loc[:, ["MX", "CA", "SA", "RU"]]
m.rolling(7, min_periods=1).mean().plot()
\end{lstlisting}
}

Figure \ref{fig:mexico_mobility_weekday} presents the information as a percentage. The percentage is computed using a baseline period, which corresponds to the 13 weeks previous to the event of interest. The baseline statistics can be computed using different procedures; in this contribution, two are described; one using the weekday and using a clustering algorithm, particularly k-means. The \emph{text models} library has two classes; one computes the percentage using weekday information, namely MobilityWeekday, and the other using a clustering algorithm, i.e., MobilityCluster. The following code lines compute the percentage using the weekday information; the code is similar to the one used to produce Figure \ref{fig:mexico_mobility_weekday}. The first sixth lines compute Mexico's mobility, and then the seventh and eighth lines transform the data into a matrix where the columns are the weeks, and each row is the different day of the week.

{\small
\begin{lstlisting}[language=Python]
import pandas as pd
from text_models import MobilityWeekday
end=dict(year=2020, month=2, day=17)
start=dict(year=2021, month=1, day=10)
mob = MobilityWeekday(start, end=end)
d = mob.overall(pandas=True).loc[:, "MX"]
d = {c.date(): v.to_numpy() 
     for c, v in d.resample("W")}
pd.DataFrame(d).boxplot(rot=90)
\end{lstlisting}
}

\subsection*{Text}

The other studies that can be performed with the library are based on tokens and their frequency per day segmented by language and country. We are in the position to replicate Figure \ref{fig:word-cloud-us} that corresponds to the word cloud produced with the tokens retrieved from the United States in English on February 14, 2020. The first line of the following code imports the Vocabulary class, and the third line instantiates the class specifying the language and country of interest. 

{\small
\begin{lstlisting}[language=Python]
from text_models import Vocabulary
day = dict(year=2020, month=2, day=14)
voc = Vocabulary(day, lang="En",
                 country="US")
\end{lstlisting}
}

The tokens used to create the word cloud are obtained after removing the emojis and frequent words. 

{\small
\begin{lstlisting}[language=Python]
voc.remove_emojis()
voc.remove(voc.common_words())
\end{lstlisting}
}

The word cloud is created using the library {\em WordCloud}. The first two lines import the word cloud library, and the library is used to produce the plot. The third line then produces the word cloud, and the fourth produces the figure; the last is just an aesthetic instruction of the plot library.

{\small
\begin{lstlisting}[language=Python]
from wordcloud import WordCloud as WC
from matplotlib import pylab as plt
wc = WC().generate_from_frequencies(voc)
plt.imshow(wc)
plt.axis("off")
\end{lstlisting}
}

As shown in the previous word cloud, the most frequent tokens are related to Valentines' day. A procedure to retrieve other topics that occurred on this day is to remove the previous years' frequent words, as depicted in Figure \ref{fig:word-cloud-us-2}.

{\small
\begin{lstlisting}[language=Python]
day_words = voc.day_words()
cw = day_words.common_words(quantile=0.90)
voc.remove(cw, bigrams=False)
\end{lstlisting}
}

The word cloud is created using a similar procedure, being the only difference the tokens in the given to the class.

{\small
\begin{lstlisting}[language=Python]
wc = WC().generate_from_frequencies(voc)
plt.imshow(wc)
plt.axis("off")
\end{lstlisting}
}

Figure \ref{fig:similarity} is created, taking randomly 180 days, from January 1, 2019, to November 1, 2021. We use different methods from text\_models library; namely, date\_range to create the period, and Vocabulary.available\_dates to obtain the first 180 dates available for all the countries with the specified languages. 

{\small
\begin{lstlisting}[language=Python]
from text_models.utils import date_range
import random

init = dict(year=2019, month=1, day=1)
end = dict(year=2021, month=11, day=1)
dates = date_range(init, end)
random.shuffle(dates)
countries = ['MX', 'CO', 'ES', 'AR',
             'PE', 'VE', 'CL', 'EC',
             'GT', 'CU', 'BO', 'DO',
             'HN', 'PY', 'SV', 'NI',
             'CR', 'PA', 'UY']
avail = Vocabulary.available_dates
dates = avail(dates, n=180,
              countries=countries,
              lang="Es")
\end{lstlisting}
}

Once the available dates are obtained, we can retrieve the tokens on the specified dates for the Spanish-speaking countries. 

{\small
\begin{lstlisting}[language=Python]
vocs = [Vocabulary(dates, lang="Es",
                   country=c)
        for c in countries]
\end{lstlisting}
}

Once the vocabulary is retrieved, it can be used to compute the similarity between countries. The similarity is obtained using Jaccard on the most common words and bigrams. We decided to use the same number of words for all countries; this number was fixed to 10\% of the country's number of words and bigrams with the smallest vocabulary. This number is obtained in the first two lines. The rest of the lines calculate the most common words and bigrams for each country.  

{\small
\begin{lstlisting}[language=Python]
_min = min([len(x.voc) for x in vocs])
_min = int(_min * .1)
tokens = [x.voc.most_common(_min) 
          for x in vocs]
tokens = [set(map(lambda x: x[0], i)) 
          for i in tokens]
\end{lstlisting}
}

The Jaccard similarity matrix is defined in set operations; the variable tokens is a list of sets, each one corresponding to a different country. Then there are two nested loops, each one iterating for the country sets. 

{\small
\begin{lstlisting}[language=Python]
X = [[len(p & t) / len( p | t)
      for t in tokens] for p in tokens]
\end{lstlisting}
}

Each row of the Jaccard similarity matrix can be used as the country signature. To depict this signature in a plane, we transformed it using Principal Component Analysis (PCA). The following code transforms the matrix into a matrix with two columns. 

{\small
\begin{lstlisting}[language=Python]
from sklearn.decomposition import PCA
X = PCA(n_components=2).fit_transform(X)
\end{lstlisting}
}

Given that a two-dimensional vector represents each country, one can plot them in a plane using a scatter plot. The second line of the following code plots the vectors in a plane; meanwhile, the loop sets the country code close to the point. 

{\small
\begin{lstlisting}[language=Python]
from matplotlib import pylab as plt
plt.plot(X[:, 0], X[:, 1], "o")
for l, x in zip(countries, X):
  plt.annotate(l, x)
\end{lstlisting}
}

% \verb+\printcredits+ command is used after appendix sections to list author credit taxonomy contribution roles tagged using \verb+\credit+ in frontmatter.

\section{Computer code availability}

\begin{itemize}
    \item Name of code: text\_models
    \item Developer: Mario Graff (mgraffg@ieee.org)
    \item Year first available: 2020
    \item Hardware required: Any platform where Python is available
    \item Program language: Python 3
    \item Access to code: \url{https://github.com/INGEOTEC/text_models}
\end{itemize}
%\bibliographystyle{cas-model2-names}


%\bibliography{ingeotec.bib}


% \bio{figs/pic1}


\end{document}